\documentclass[11pt]{article}
\usepackage{UF_FRED_paper_style}
\usepackage{lipsum}
\title{Actionable Phrase Detection using NLP}

\author{Adit Magotra\\
    \href{mailto:aditmagotra@gmail.com}{\texttt{aditmagotra@gmail.com}} 
    }
    
\date{September 2022}

\begin{document}
{\setstretch{.8}
\maketitle
\begin{abstract}

In this paper, the aim is to explore if Actionables (Task phrases) can be extracted from raw text using Linguistic filters designed from scratch. These filters are specially catered to identifying actionable text using Transfer Learning as the lead role. Actionable Detection can be used in detecting emergency tasks during a crisis, Instruction accuracy for First aid and can also be used to make productivity tools like automatic ToDo list generators from conferences. To accomplish this, we use the Enron Email Dataset and apply our Linguistic filters on the cleaned textual data. We then use Transfer Learning with the Universal Sentence Encoder to train a model to classify whether a given string of raw text is an actionable or not. 

\end{abstract}
}


\section{Introduction}

Actionable sentences are terms that, in the most basic sense, imply the necessity of taking a specific action. In Linguistic terms, they are steps to achieve an operation, often through the usage of action verbs. 

For example, the sentence, “Get your homework finished by tomorrow” qualifies as actionable since it demands a specific action (In this case, finishing homework) to be taken. In contrast, a simple sentence such as, “I like to play the guitar” does not qualify as an actionable phrase since it simply states a personal choice of the person instead of demanding a task to be finished.

They have practically limitless applications, from task detection for boosting productivity to urgent action detection in emergency situations like surgery and natural disasters. 

The main verb in the sentence that indicates a task is termed as the action verb. There are several methods for identifying action verbs in a sentence, a popular method being POS-Tagging. POS taggers are trained with long newswire text and not with short, sketched sentences like task labels, so to improve further the accuracy we performed manual amendment of some tags (especially verbs instead of nouns)[1]. The ten most frequent structures are shown in Table 1.0.

\begin{table}[]
\begin{tabular}{ccccc}
\hline
Group Tag & POS Tags               & Occurrence & Percentage & Unique First Words \\ \hline
verb      & VB, VBD, VBG, VBP, VBZ & 1,173      & 73.63      & 70                 \\
noun      & NN, NNS, NNP, NNPS     & 322        & 20.21      & 65                 \\
adjective & JJ, JJR, JJS           & 27         & 1.69       & 13                 \\
adverb    & RB, RBR, RBS           & 27         & 1.69       & 4                  \\
pronoun   & PRP, PRP\$             & 7          & 0.44       & 2                  \\
other     & \multicolumn{1}{l}{}   & 37         & 2.32       & 11                 \\
total     & \multicolumn{1}{l}{}   & 1,593      & 100        & 165                \\ \hline
\end{tabular}
\caption{\label{Table 1.0}Grouping of POS tags employed in analysis}
\end{table}
\section{Dataset}
There is currently no dataset that is tailored to the recognition of task phrases. An algorithm was then used to create our own dataset, consisting of multiple processes that combined linguistic and mechanical methods to weed out statements that are very likely to be useful. 

However, to make these adjustments, a baseline dataset of similar capability is required which can then be modified into a full-fledged actionable detection dataset. We use the Enron Email Dataset [2] for this purpose.\vspace{\baselineskip}
\noindent \\
\textbf{Baseline Dataset} \\
The Enron email dataset contains approximately 500,000 emails generated by employees of the Enron Corporation. It was obtained by the Federal Energy Regulatory Commission during its investigation of Enron's collapse.

It serves as a perfect corpus for our purpose because emails sent by an organization’s employees to each other are very likely to contain tasks assigned to another employee and the model is more likely to figure out an actionable correlation from such a dataset. It is also a fairly large dataset containing 500,00+ emails providing our model with a plethora of features to train on.

\vspace{\baselineskip}

\begin{figure}[H]
    \centering
        \includegraphics[scale=.45]{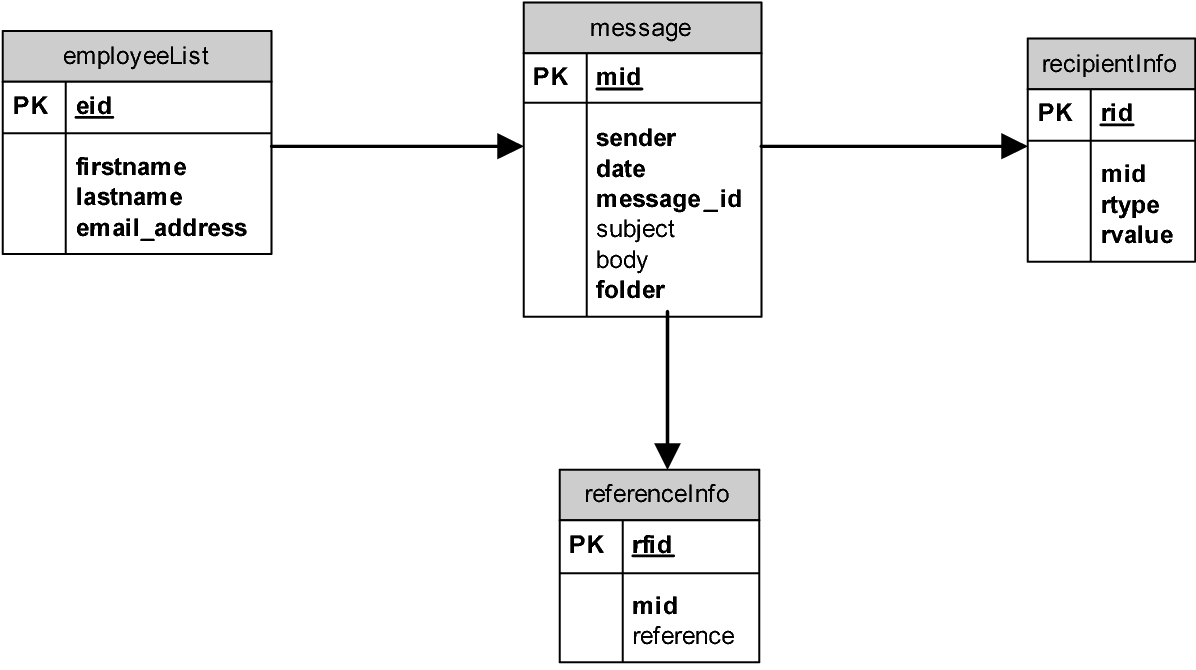}
    \caption{Enron Distribution [6]}
    \label{fig:1}
\end{figure}

\vspace{\baselineskip}
\noindent \\
\textbf{Transformation Approach} \\
To classify whether or not a sentence is an actionable phrase, several filters were tested and experimented with to find out which ones correlate the most with task phrases and can be used to train a classification model.

Table 2.0 shows all the proposed filters that were used in the extraction of sentences from the dataset.

\vspace{\baselineskip}

\begin{figure}[H]
    \centering
        \includegraphics[scale=.85]{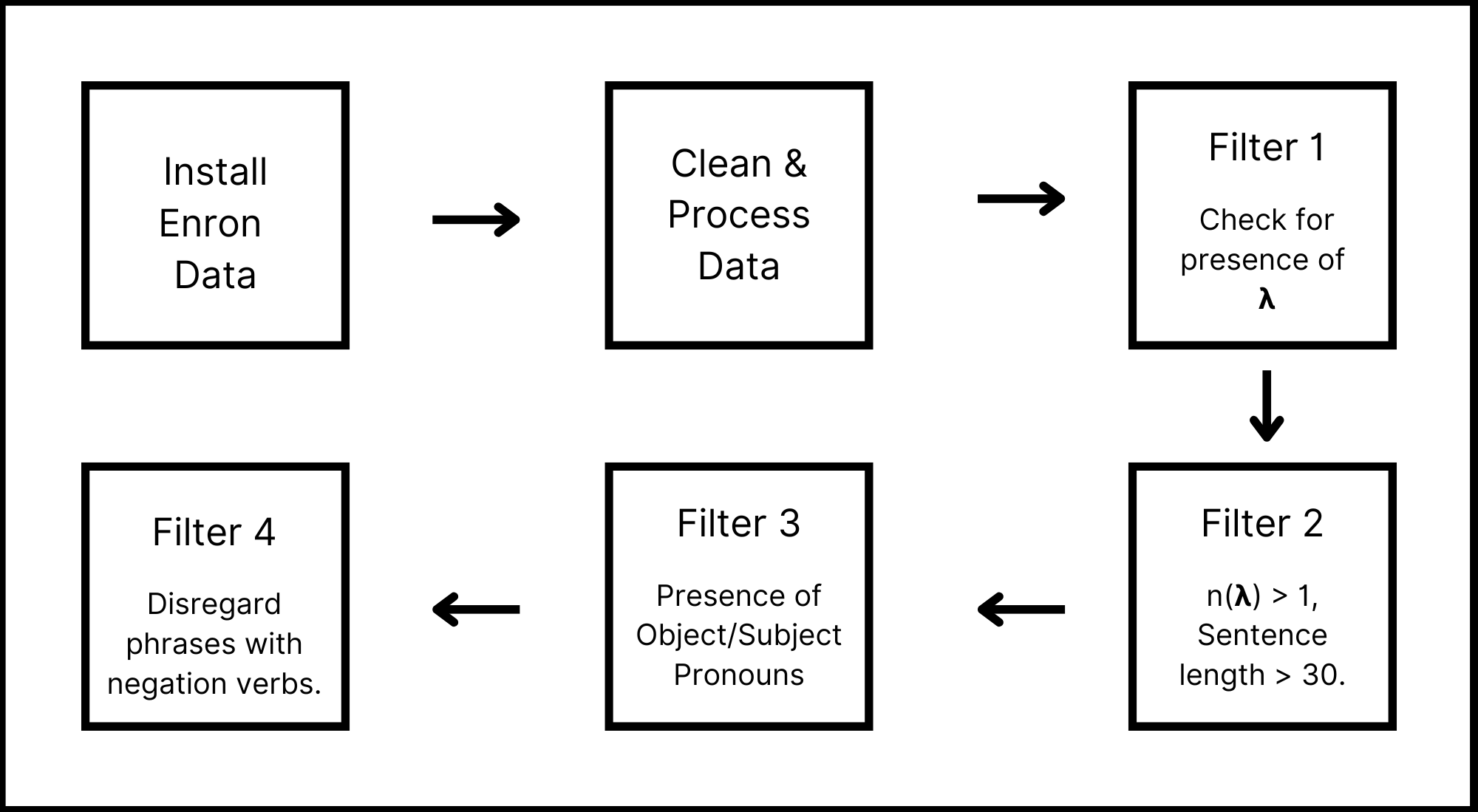}
    \caption{Flowchart depicting steps and filters ( $\lambda$  depicts ‘Action Verbs’ )}
    \label{fig:1}
\end{figure}

\vspace{\baselineskip}
\noindent \\
\textbf{Filter 1}\\
As expected, the metric that showed the highest interrelation with the result was the presence of action verbs in the sentence. 

The filtration requests modality-rich input for crossmodal, cross-situational learning of word-object and word-action mappings. Some corpora explored included the fully-fledged Action Verb Corpus [3]. 

However, for experimentation purposes, a concentrated .txt file of a few action verbs was used. Table 2.1 depicts some of the action verbs used to filter out phrases from the dataset.

\vspace{\baselineskip}

\begin{table}[h]
\centering
\begin{tabular}{|c|}
\hline
Arrive | Build | Close | Define | Formulate | Propose | Kickstart … \\ \hline
\end{tabular}
\caption{\label{Table 1.0}Sample Action Verbs}
\end{table}

\vspace{\baselineskip}
\noindent \\

\textbf{Filter 2}\\
Checking for the length of the sentence is an important aspect as it gives an idea of the ratio of the no. of action words to the total length of the sentence. If a single action verb is used in a very long sentence, it might lose its value.\vspace{\baselineskip}
\noindent \\

\textbf{Filter 3}\\
An object pronoun is a personal pronoun that is used typically as a grammatical object: the direct or indirect object of a verb, or the object of a preposition while any noun performing the main action in the sentence, is a subject and is categorized as subjective case

\vspace{\baselineskip}

\begin{table}[h]
\centering
\begin{tabular}{|l|c|}
\hline
Object Pronouns  & me | her | him | us | them     \\ \hline
Subject Pronouns & I | we | you | he | she | they \\ \hline
\end{tabular}
\end{table}

The subject noun, in this case, acts as a direct reference to the actionable in the sentence while the object pronoun helps in amplifying the signal for a more accurate response.\vspace{\baselineskip}
\noindent \\

\textbf{Filter 4}\\
The final filter involves disregarding sentences with negation verbs such as “shouldn’t”, “couldn’t” and “wouldn’t”. Usually, a sentence containing these terms is not likely to be actionable because a negation would usually denote the process of an action not being taken. Hence, including sentences with negation words might be detrimental for the model because if a sentence contains an action verb and a negation verb, the model might count it as an actionable

\section{Modelling}

First, a training set and testing set had to be created in order to train machine learning models [4]. Therefore, we randomly divided the dataset into 20\% for testing and 80\% for training. This was done programmatically. Randomization determined whether a phrase was assigned to the training or testing set.\vspace{\baselineskip}
\noindent \\
\textbf{Tree-Based Modelling}\\
Several ‘classic’ machine learning models were tested and experimented-with, where tree-based models took the lead. However, for training classic models, the text had to first be vectorised using TFIDF and and tokenized using the Natural Language Toolkit as a Python library.The approach produced tangible results. However, the F1 score could be increased by a major chunk by using state-of-the-art NLP methods.

\vspace{\baselineskip}

\begin{figure}[H]
    \centering
        \includegraphics[scale=.85]{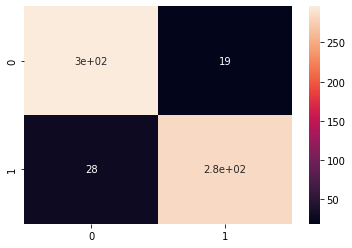}
    \caption{Random Forest Result ( $\lambda$  depicts ‘Action Verbs’ )}
    \label{fig:1}
\end{figure}

\vspace{\baselineskip}
\noindent \\
\textbf{Universal Sentence Encoder}\\

The Universal Sentence Encoder makes getting sentence level embeddings as easy as it has historically been to lookup the embeddings for individual words. The sentence embeddings can then be trivially used to compute sentence level meaning similarity as well as to enable better performance on downstream classification tasks using less supervised training data.

\vspace{\baselineskip}

\begin{figure}[H]
    \centering
        \includegraphics[scale=.3]{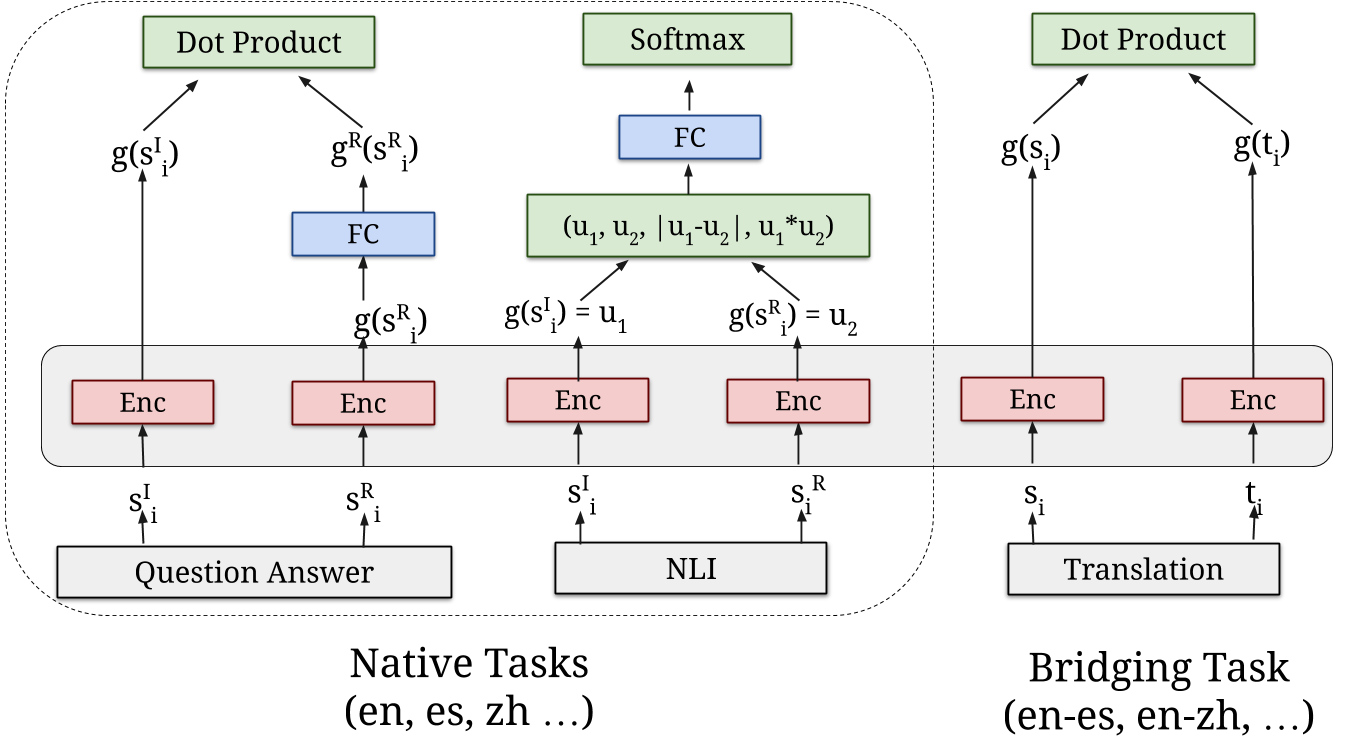}
    \caption{Universal Sentence Encoder [5]}
    \label{fig:1}
\end{figure}

The Univeral Sentence encoder was setup using Tensorflow Hub with parameters that disallowed retraining of the model. It was a part of a larger architecture which allowed raw text to be given as input and returned a probability prediction of the possibility of a sentence being an actionable.

A Layered Sequential model was created with the Universal Sentence encoder as it’s first (input) layer. A dropout layer was added which ignores weights of  20\% of random neurons to prevent overfitting. A Dense layer was added with 64 units for feature extraction, activated by ReLU. It was followed by another Sigmoid activated Dense layer with 1 unit to concentrate the output into a single probability value. 
\vspace{\baselineskip}

\begin{figure}[H]
    \centering
        \includegraphics[scale=.6]{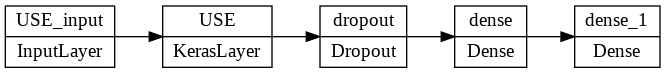}
    \caption{Sequential Model Workflow}
    \label{fig:1}
\end{figure}

The model was then trained with 10 epochs using the Adam optimizer and Binary CrossEntropy loss metric. 

Then, the probability is matched with a median threshold value function which converts the probability to a binary numeral which denotes if the phrase is an actionable or not.

\section{Results}

The metrics added for the USE model were Accuracy, F1 Score, Precision and Recall. 

\vspace{\baselineskip}
\vspace{\baselineskip}

\begin{table}[h]
\centering
\begin{tabular}{|l|l|}
\hline
Accuracy  & \begin{tabular}[c]{@{}l@{}}Training: 0.923\\ Validation: 0.914\\ Testing: 0.908\end{tabular} \\ \hline
F1 Score  & 0.929                                                                                        \\ \hline
Precision & 0.884                                                                                        \\ \hline
Recall    & 0.906                                                                                        \\ \hline
\end{tabular}
\caption{Metrics and Values}
\end{table}

\vspace{\baselineskip}
\vspace{\baselineskip}

\noindent
The training and testing accuracy, loss was captured for both testing and training sets and plotted using matplotlib.

\begin{figure}[h]
    \centering
    \subfloat[\centering label 1]{{\includegraphics[width=6.5cm]{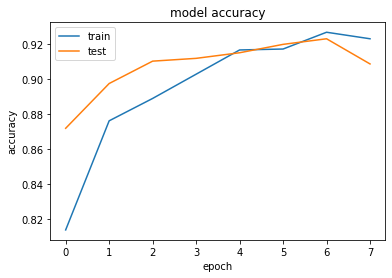} }}%
    \qquad
    \subfloat[\centering label 2]{{\includegraphics[width=6.5cm]{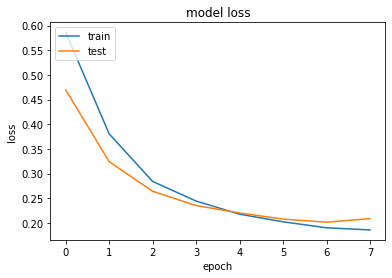} }}%
    \caption{2 Figures side by side}%
    \label{fig:example}%
\end{figure}


\medskip

\newpage

\section{References}

\vspace{\baselineskip}

\noindent
[1] Müter, L., Deoskar, T., Mathijssen, M., Brinkkemper, S., Dalpiaz, F. (2019). Refinement of User Stories into Backlog Items: Linguistic Structure and Action Verbs. In: Knauss, E., Goedicke, M. (eds) Requirements Engineering: Foundation for Software Quality. REFSQ 2019. Lecture Notes in Computer Science(), vol 11412. Springer, Cham.\\ https://doi.org/10.1007/978-3-030-15538-4\_7  

\vspace{\baselineskip}
\vspace{\baselineskip}

\noindent
[2] Enron Corp \& Cohen, W. W. (2015) Enron Email Dataset. United States Federal Energy Regulatory Commissioniler, comp [Philadelphia, PA: William W. Cohen, MLD, CMU] [Software, E-Resource] Retrieved from the Library of Congress\\ https://www.loc.gov/item/2018487913/. 

\vspace{\baselineskip}
\vspace{\baselineskip}

\noindent
[3] Gross (Schreitter), Stephanie \& Hirschmanner, Matthias \& Krenn, Brigitte \& Neubarth, Friedrich \& Zillich, Michael. (2018). Action Verb Corpus.\\ https://aclanthology.org/L18-1338.pdf 

\vspace{\baselineskip}
\vspace{\baselineskip}

\noindent
[4] Shah, T. (2020, July 10). About Train, Validation and Test Sets in Machine Learning - Medium.\\ https://towardsdatascience.com/train-validation-and-test-sets-72cb40cba9e7 

\vspace{\baselineskip}
\vspace{\baselineskip}

\noindent
[5] Yinfei Yang and Amin Ahmad (2019, July 12). Multilingual Universal Sentence Encoder for Semantic Retrieval.\\ https://ai.googleblog.com/2019/07/multilingual-universal-sentence-encoder.html 

\vspace{\baselineskip}
\vspace{\baselineskip}

\noindent
[6] Shetty, J., \& Adibi, J. (2004). The Enron Email Dataset Database Schema and Brief Statistical Report.\\ https://foreverdata.org/1009HOLD/Enron\_Dataset\_Report.pdf 

\end{document}